\documentclass[a4paper,UKenglish]{lipics-v2016}

\usepackage{booktabs} 

\usepackage[ruled]{algorithm2e} 

\usepackage{amsmath}
\usepackage{amssymb}
\usepackage{lmodern}
\usepackage{array}

\usepackage{subcaption}
\usepackage{pgfplots}
\pgfplotsset{width=10cm,compat=1.9}

\usepackage{multirow}
\usepackage{tikz}
\usetikzlibrary{matrix}

\title{Evaluating approaches for supervised semantic labeling}
\titlerunning{Semantic labeling} 

\author[1]{Natalia R\"ummele\footnote{Work accomplished at Data61, CSIRO.}}
\author[2]{Yuriy Tyshetskiy}
\author[3]{Alex Collins}
\affil[1]{Siemens, Germany\\
  \texttt{nataliia.ruemmele@siemens.com}}
\affil[2]{Data61, CSIRO, Australia\\
  \texttt{yuriy.tyshetskiy@data61.csiro.au}}
\affil[3]{Data61, CSIRO, Australia\\
  \texttt{alex.collins@data61.csiro.au}}
\authorrunning{N. R\"ummele, Y. Tyshetskiy, A. Collins} 

		
\begin{document}

\maketitle

\begin{abstract}
Relational data sources are still one of the most popular ways to store enterprise or Web data, however, the issue with relational schema is the lack of a well-defined semantic description.
A common ontology provides a way to represent the meaning of a relational schema and can facilitate the integration of heterogeneous data sources within a domain.
Semantic labeling is achieved by mapping attributes from the data sources to the classes and properties in the ontology.
We formulate this problem as a multi-class classification problem where previously labeled data sources are used to learn rules for labeling new data sources.
The majority of existing approaches for semantic labeling have focused on data integration challenges such as naming conflicts and semantic heterogeneity.
In addition, machine learning approaches typically have issues around class imbalance, lack of labeled instances and relative importance of attributes.
To address these issues, we develop a new machine learning model with engineered features as well as two deep learning models which do not require extensive feature engineering.
We evaluate our new approaches with the state-of-the-art.
\end{abstract}

\section{Introduction}\label{sec:intro}

An important problem in database research is determining how to combine multiple data sources that are described by different (heterogeneous) schemata~\cite{Doan:di}.
The outcome of such a process is expected to be a uniform integrated view across these data sources.
Relational data sources are still one of the most popular ways to store enterprise or Web data~\cite{Spanos:semweb}.
However, the relational schema lacks a well-defined semantic description.
To define the semantics of data, we can introduce an ontology~\cite{Spanos:semweb}.
Now our goal is to map attributes from relational data sources to classes and properties in an ontology.
We refer to this problem as \emph{semantic labeling}.

Semantic labeling plays an important role in data integration~\cite{Doan:di,Pham:semantic}, augmenting existing knowledge bases~\cite{Limaye:Annotating,Ritze:matching,Ritze:HTML,Venetis:Recovering} or mapping relational sources to ontologies~\cite{Pinkel:rodi,Taheriyan:semantics}.
Various approaches to automate semantic labeling have been developed, including DSL~\cite{Pham:semantic} and T2K~\cite{Ritze:matching}. Typically automated semantic labeling techniques encounter several problems.
Firstly, there can be naming conflicts~\cite{Pinkel:rodi}, including those cases where users represent the same data in different ways.
Secondly, semantically different attributes might have syntactically similar content, for example, birth date versus date of death.
Thirdly, there are a considerable number of attributes which do not have any corresponding property in the ontology, either by accident or on purpose.
The majority of existing systems focus on the first two problems, but do not consider the third problem during evaluation~\cite{Ritze:HTML,Pham:semantic}.

To address the challenges of automated semantic labeling, we formulate this task as a supervised classification problem.
A set of semantic labels known to the classifier is specified at training time, e.g., from the provided domain ontology.
We also introduce a special class of attributes, called \emph{unknown}.
The purpose of the unknown class is to capture attributes which will not be mapped to the ontology.
The training data for the classifier will thus consist of source attributes (name and content) and their semantic labels provided by the user, including the \emph{unknown} labels.
Since manually assigning labels to attributes is a costly operation, a lack of training data is a common problem for semantic labeling systems.
Existing systems~\cite{Pham:semantic,Ritze:matching,Venetis:Recovering} use knowledge transfer techniques to overcome this issue.
Instead, we introduce a sampling method similar to bagging for ensemble models~\cite{Breiman:bagging}.

The bagging technique allows us to generate multiple training instances from the user-labeled attributes, thus overcoming the lack of labeled training data.
It also allows us to overcome the common issue of class imbalance, when some semantic labels have more support than others among the attributes.
We can achieve this by re-balancing the training data via preferential bagging from minority class attributes.

The main contributions of this paper are:
\begin{enumerate}
\item We introduce a bagging approach to handle class imbalance and the lack of training data by drawing random subsamples from values of an attribute.
 This approach can achieve meaningful diversity in the training data and can increase the number of training instances for under-represented semantic labels.
\item We address the issue of ``unwanted'' attributes, i.e., attributes which do not get mapped to any element in the ontology.
In cases where we have a sufficient amount of training data, our models can achieve over 80\% Mean Reciprocal Rank (MRR) on two sets of data sources from our benchmark.
\item We construct a classification model \emph{DINT} with hand-engineered semantic labeling features to implement the above.
In addition, we design two deep learning models \emph{CNN} and \emph{MLP} which use very simple features, such as normalized character frequencies and padded character sequences extracted from raw values of data attributes.
\item We construct a benchmark with a common evaluation strategy to compare different approaches for supervised semantic labeling.
Our benchmark includes such models as \emph{DINT}, \emph{CNN}, \emph{MLP} and the state-of-the-art \emph{DSL}~\cite{Pham:semantic}, and 5 sets of data sources from different domains.
We show that each approach has its strengths and shortcomings, and choosing a particular semantic labeling system depends on the use case.
We have released the implementation of the benchmark under an open source license~\footnote{\href{http://github.com/NICTA/serene-benchmark}{http://github.com/NICTA/serene-benchmark}}.
This benchmark can be easily extended to include other models and datasets, and can be used to choose the most appropriate model for a given use case.
\end{enumerate}

\section{Problem}\label{sec:problem}

\begin{figure}%
	\centering
	\includegraphics[width=0.9\columnwidth]{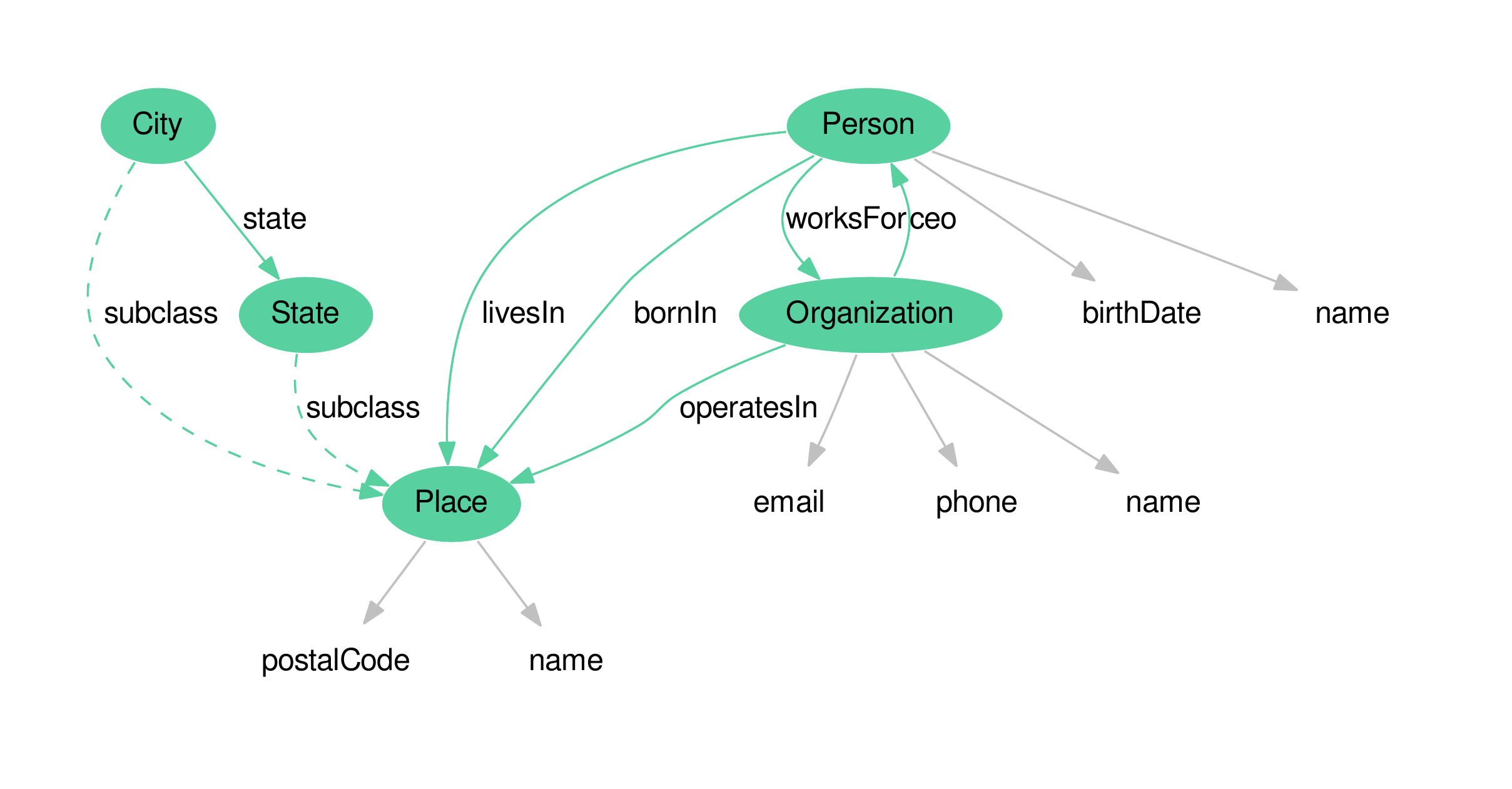}%
	\caption{Example ontology.}%
	\label{fig:ontology}%
\end{figure}

We illustrate the semantic labeling problem using a simple domain ontology shown in Fig.~\ref{fig:ontology}.
Assume we have three data sources ``personal-info'', ``businessInfo'' and ``Employees'' (see Tab.~\ref{tab:relexamples}) whose attributes we choose to label according to the example ontology (Fig.~\ref{fig:ontology}).
We define a \emph{semantic label} as a tuple consisting of a domain class and its property.
For example, attribute \emph{name} in the source ``personal-info'' (see Tab.~\ref{tab:s1}) is labeled with \emph{(Person,name)}.
Note that semantic labels are fixed by the ontology.

The task of semantic labeling is defined as automatically assigning semantic labels to attributes in a data source.
In the case of supervised semantic labeling, we use existing known semantic labels for data sources to improve the performance when assigning semantic labels to new sources.
For example, assume we are given sources ``personal-info'' and ``businessInfo'' with the correct semantic labels, the system should then automatically assign labels to attributes in the source ``Employees''.

To build such a system, we cannot just rely on the names of the columns.
For example, columns \emph{name} in~(\ref{tab:s1}), \emph{ceo} in~(\ref{tab:s3})
and \emph{employee} in~(\ref{tab:s4}) all refer to the same property \emph{(Person,name)}.
Using just values of the columns is also problematic.
For example, in~(\ref{tab:s1}) acronyms are used for states, while in~(\ref{tab:s3}) state names are fully written.
Furthermore, values can overlap for semantically heterogeneous columns like for \emph{founded} in~(\ref{tab:s3}) and \emph{birthDate} in~(\ref{tab:s1}). 

\begin{table*}[t]\small
  \centering
  \caption{Example relational data sources with semantic labels.}
	\begin{subtable}{\textwidth}
	\centering
		\begin{tabular}{ cccccc}
		\cmidrule{2-6}
		& \textbf{name} & \textbf{birthDate} & \textbf{city} & \textbf{state} & \textbf{workplace}\\
		\cmidrule{2-6}
		& Neil & 21-05-1916 & Waterloo & NSW & CSIRO\\
		& Mary & 07-12-1990 & Eveleigh & NSW & CSIRO\\
		& Henry & 15-03-2000 & Redfern & NSW & Data61\\
		\cmidrule{2-6}
		\emph{Semantic} & \emph{(Person,} & \emph{(Person,} & \emph{(City,} & \emph{(State,} & \emph{(Organization,} \\
		\emph{labels} & \emph{name)} & \emph{birthDate)}  & \emph{name)} & \emph{name)} & \emph{name)} \\
		\cmidrule{2-6}
		\end{tabular}
	\caption{personal-info}\label{tab:s1}
	\end{subtable}
	\qquad%
	\begin{subtable}{\textwidth}
	\centering
		\begin{tabular}{ccc}
		\toprule
		\textbf{employer} & \textbf{employee} & \textbf{DOB}\\
		\midrule
		CSIRO & Neil & 05/21/1916\\
		Data61 & Mary & 12/07/1990\\
		NICTA & Henry & 03/15/2000\\
		\midrule
		\emph{(Organization,} & \emph{(Person,} & \emph{(Person,} \\
		\emph{name)} &  \emph{name)} & \emph{birthDate)} \\
		\bottomrule
		\end{tabular}
	\caption{Employees}\label{tab:s4}
	\end{subtable}
  	\qquad%
	\begin{subtable}{\textwidth}
	\centering
		\begin{tabular}{ ccccc}
		\cmidrule{2-5}
		& \textbf{company} & \textbf{ceo} & \textbf{state} & \textbf{founded}\\
		\cmidrule{2-5}
		& CSIRO & Larry Marshall & Australian Capital Territory & 21-05-1916\\
		& Data61 & Adrian Turner  & New South Wales & 12-07-2016\\
		& NICTA & Hugh Durrant & New South Wales & 15-03-2002\\
		\cmidrule{2-5}
		\emph{Semantic} & \emph{(Organization,} & \multirow{2}{*}{\emph{(Person,name)}} & \multirow{2}{*}{\emph{(State,name)}} &  \multirow{2}{*}{\centering \emph{unknown}}\\
		\emph{labels} & \emph{name)} &   &   &  \\
		\cmidrule{2-5}
		\end{tabular}
	\caption{businessInfo}\label{tab:s3}
	\end{subtable}
	\label{tab:relexamples}
\end{table*}

We can also have attributes that are not mapped to any property in the ontology.
There might be two reasons for their existence:
(1) we are not interested in the content of an attribute and want to discard it from any future analysis;
(2) we might have overlooked an attribute by not designing the ontology accurately.
We do not differentiate between these two cases and mark all such attributes as \emph{unknown class}, for example, \emph{founded} in~(\ref{tab:s3}).
The presence of unknown class makes the task of semantic labeling more complicated.
Establishing approaches to efficiently handle such attributes is crucial since in many real-world scenarios relational data sources (either HTML tables~\cite{Ritze:matching} or domain specific data~\cite{Pham:semantic}) contain a considerable number of such attributes.

Machine learning techniques proved to be efficient in building predictive models on noisy and messy data.
Yet to apply these techniques we need to represent source attributes as \emph{feature vectors}, with semantic labels (classes) attached to these vectors.
In Table~\ref{tab:feature} we show such representation for the source Employees.
We have explicitly shown only 4 possible features, for simplicity.
However, the actual size of a feature vector can be arbitrary long, and the process of designing its components is known as \emph{feature engineering}.
In the next section we will discuss the features used in the semantic labeling system.

\begin{table*}[t]\small
  \centering
  \caption{Feature vectors for data source Employees.}\label{tab:feature}
	\begin{tabular}{ c|ccccc|c}
	\toprule
	\multirow{3}{*}{\textbf{attribute}} & \multicolumn{5}{c|}{\textbf{feature vector}} & \multirow{3}{*}{\textbf{class}} \\
	& entropy & mean & \multirow{2}{*}{$\ldots$} & ratio & ratio unique & \\
	&  & string length &  & alpha chars & values & \\
	\midrule
	employer & $1.001$  & $5.333$ & $\ldots$ & $0.875$ & 1 & (Organization, name)\\
	employee & $1.461$ & $13.333$ & $\ldots$ & $0.925$  & 1 & (Person, name)\\
	DOB & $0.69$ & $10$  & $\ldots$ & $0$ & 1 & (Person, birthDate)\\
	\bottomrule
	\end{tabular}
\end{table*}

\section{Approaches}\label{sec:approach}

In this section we describe classifiers for the semantic labeling problem used for evaluation.
We also discuss approaches to the problem of unknown attributes and lack of training data.

Once we have a set of labeled data sources, we construct feature vectors for all attributes in this set and mark them as representatives of a class corresponding to their semantic labels.
The constructed set of (feature vector, class label) pairs is then used to train a classifier.
We consider several approaches, divided into 3 major groups: \emph{DINT}, Deep Learning and the state-of-the-art \emph{DSL}.
Each approach trains a multi-class classification model that produces, at the prediction stage, a list of class probabilities for an attribute in a new source.
The class with the highest predicted probability is then assigned to the attribute at the decision stage.

\subsubsection{DINT}\label{sec:dint}
In our first approach \emph{DINT} (Data INTegrator) we hand-engineer 26 features, which include characteristics such as number of whitespaces and other special characters, statistics of values in the column (e.g, mean/ max/ min string length and numeric statistics) and many more.
The complete list of features is available in the open source benchmark repository~\footnote{\href{http://github.com/NICTA/serene-benchmark/tree/master/doc}{serene-benchmark}}.
One of the important features characterising information content of an attribute is Shannon's entropy of the attribute's concatenated rows.
Shannon's entropy (or information entropy~\cite{Manning:Introduction}) of a string $X$ is defined as
$H(X) = -\sum_{i}{p_i \log_{2}p_i},$ where $p_i$ is the probability of a character, whose index in character vocabulary is $i$, to appear in $X$, and the summation ranges over all characters in the vocabulary.
To evaluate $p_i$ in Shannon's entropy, we evaluate normalized character frequency distribution \emph{chardist} of an attribute, as character counts in concatenated rows of the attribute, normalized by the total length of the concatenated rows.
The vocabulary of all characters consists of 100 printable characters (including $\backslash$n).
Finally, we also add the 100-dimensional vector of $p_i$ to the attribute feature vector.

In addition to the above features, which can be directly calculated from attribute values, we compute mean cosine similarity of attribute character distribution with character distributions of all class instances.
This adds as many additional scalar features to the full attribute feature vector as there are classes in the training data.

One can expect that names of the attributes should also contain useful information to determine their semantic types, in addition to the information provided by attribute values.
To extract features from attribute names, we compute string similarity metrics: minimum edit distance, two WordNet based similarity measures such as JCN~\cite{Jiang:Semantic} and LIN ~\cite{Lin:Information}, and $k$-nearest neighbors using Needle-Wunsch distance~\cite{Needleman:General}.
The minimum edit distance between two strings $s_1$ and $s_2$ is the minimum number of edit operations, such as insertion, deletion, substitution, which are required to transform one string into another~\cite{Manning:Introduction}.
We compute the similarity between attribute name and all class instances in the training data.
The number of thus extracted features depends on the number of semantic labels in the training data.

We choose to train a Random Forest~\cite{Breiman:RF} (RF) on this set of features.
RF is quite robust on noisy data, works well even with correlated features, and easily captures complex nonlinear relationships between features and target.
Additionally, RF classifiers require little hyperparameter tuning, and hence they usually work straight ``out of the box'', which makes them a convenient yet versatile classifier to use.

\subsubsection{Deep Learning}\label{sec:dl}

Deep learning has gained much popularity due to its tremendous impact in such areas as speech recognition, object recognition, and machine translation~\cite{Lecun:deep}.
One of the biggest advantages of deep learning is the ability to process data in its raw form and to discover the representation needed for classification, assisting with the feature engineering step.

Broadly speaking, deep learning is an overarching term for artificial neural networks, where the word ``deep'' refers to the depth of the network.
At the basic level neural networks are composed of perceptrons, or neural nodes.
There can be several layers of interconnected neural nodes;
The first layer is the input layer while the last one is the output layer.
The layers in between these two are called \emph{hidden}.
Neural nodes in each layer take as input the output of the nodes from the previous layer, perform some computation with a nonlinear activation function (e.g., tanh or RELU) and pass the result to the next layer.
There are generally no connections between nodes in the same layer.
Overall, deep learning models improve in their performance the more data they are trained on.
The exact architecture of deep learning models, i.e.,  number of layers, number of nodes in each layer, activation functions of neurons and interconnectedness between layers, all influence the performance of the trained models.

We choose two different architectures for our deep learning classifiers: (i) Multi-Layer Perceptron (\emph{MLP})~\cite{Rumelhart:learning} and (ii) Convolutional Neural Network (\emph{CNN})~\cite{Lecun:deep}.
We have experimented with different designs of the MLP and CNN networks, varying their hyperparameters that control the number of hidden layers, the numbers of nodes/filters per layer, dropout probability, etc., and found that the designs, described briefly below, work well for all the datasets in the benchmark.

The input layer of the MLP architecture takes the 101-dimensional feature vector of character frequencies $p_i$ (chardist) and Shannon entropy.
Following the input layer, MLP has 3 fully connected hidden layers with 100 nodes per layer, with tanh activations. After the 1st hidden layer, we introduced a stochastic dropout layer with dropout probability of 0.5, to prevent overfitting. Finally, the output layer of MLP (the actual classifier) is a softmax layer with the number of nodes equal to the number of semantic types (including the `unknown' type).

The CNN model takes as input the one-hot representation of an attribute's concatenated rows in character space, then embeds it to a dense 64-bit embedding, then passes this embedded "image" of the attribute through two consecutive 1d convolution layers with 100 filters per layers, followed by a 1-d max-pooling layer, a flattening layer, a dropout layer with probability of dropout 0.5, then a fully connected layer with 100 nodes, and finally a fully connected softmax output layer (the classifier) with the number of nodes equal to the number of semantic types (including the `unknown' type).

Though we cannot be sure that our final choice for the architectures is optimal, it seems to be a good trade-off between complexity of the models, required computational resources for their training, and their overall performance in semantic labeling task.
We have implemented both models using Keras library with GPU-based TensorFlow backend~\cite{Abadi:tensor}.

\subsubsection{DSL}\label{sec:dsl}

The Domain-independent Semantic Labeler (\emph{DSL}) has been proposed by Pham et al~\cite{Pham:semantic}, where 6 feature groups based on similarity metrics are constructed.
These metrics measure how attribute names and values are similar to the characteristics of other attributes.
This means that given 5 attributes in the training data (i.e., already labeled instances) with distinct semantic labels, a new attribute will be compared to representatives of each semantic label and 30 features will be calculated in total.
The considered similarity metrics are: attribute name similarity, standard Jaccard similarity for textual data and a modified version for numerical data, TF-IDF cosine similarity, distribution and histogram similarity.

Instead of building one multi-class classifier, the authors train binary classifiers separately for each semantic label.
A binary classifier for a particular semantic label is a Logistic Regression model trained on a set of similarity metrics with representatives of this label.
When predicting semantic labels for a new attribute, they combine the predictions of each classifier to produce the final vector of probabilities.
One of the distinctive properties of this approach is the ability to transfer the classification model trained in one domain to predicting semantic labels for attributes in another domain.
We denote this enhanced approach as~\emph{DSL+}.

\subsection{Bagging}\label{sec:bagging}

To train a classifier for semantic labeling, we need data sources to have many labeled attributes.
However, the costly operation of manually assigning labels to attributes, and the relative small number of columns compared to data set size, implies that lack of training data is a common problem for semantic labeling systems.
Existing systems~\cite{Pham:semantic,Ritze:matching,Venetis:Recovering} use knowledge transfer techniques to overcome this issue.
We introduce a method for increasing training sample size based on a machine learning approach known as bagging~\cite{Breiman:bagging}.

Breiman~\cite{Breiman:bagging} introduced the concept of bootstrap aggregating, also known as bagging, to construct ensembles of models to improve prediction accuracy.
The method consists in training different classifiers with bootstrapped replicas of the original dataset.
Hence, diversity is obtained with the resampling procedure by the usage of different data subsets.
At the prediction stage each individual classifier estimates an unknown instance, and a majority or weighted vote is used to infer the class.

We modify the idea of bagging for our problem.
It is clear that the semantics of columns in the table ``Employees'' (Table~
\ref{tab:s4}) will not change whether we have 3 or 1000 rows.
So, we can create several training instances for an attribute, where each instance (called a \emph{bag}) will contain a random sample (with replacement) of its content.
This procedure is governed by two parameters \emph{numBags} and \emph{bagSize}: the first parameter controls how many bags are generated per each attribute, while the latter indicates how many rows are sampled per each bag.
In such a way we address the issue of noise by increasing diversity of the training data as well as the issue of insufficient training data.

Another common problem encountered in a wide range of data mining and machine learning initiatives is \emph{class imbalance}.
Class imbalance occurs when the class instances in a dataset are not equally represented.
In such situation, building standard machine learning models will lead to poor results, since they will favor classes with large populations over the classes with small populations.
To address this issue, we have tried several resampling strategies to equalize the number of instances per each class.

\subsection{Unknown class}\label{sec:unknown}

As mentioned previously, some attributes are not mapped to any property in the ontology.
To handle this issue, we introduce one more class called \emph{unknown}.
For example, attributes which get discarded from the integration process can be marked as unknown.
This way we can help the classifier recognize such attributes in new sources.
In addition, there is another advantage of having the unknown class defined explicitly.
Consider a new attribute with an unseen semantic label, that is, a label which is not present in the training data.
Instead of picking the closest match among known semantic labels, the classifier will mark it as unknown.
The user will then need to validate the attributes that are classified as unknown.
This will ensure that the unknown class consists only of unwanted attributes.
We do not introduce another class to differentiate between unwanted attributes and unseen labels since we cannot guarantee that there is no overlap between them.
Only our \emph{DINT} and Deep Learning approaches support an unknown class.

\section{Experiments}\label{sec:experiments}

We have run all our experiments on a Dell server with 252 GiB of memory, 2 CPUs with 4 cores each, 1 Titan GPU and 1 GeForce 1080 Ti GPU.
The deep learning models have been optimized for GPUs using Tensorflow.
The benchmark for semantic labeling system is implemented in Python and is available under an open source license~\footnote{\href{http://github.com/NICTA/serene-benchmark}{http://github.com/NICTA/serene-benchmark}}.

\subsection{Datasets}

We use 5 different sets of data sources in our evaluation, labeled as: museum, city, weather, soccer~\cite{Pham:semantic} and weapons~\cite{Taheriyan:Leveraging}.
Each set corresponds to a domain with a specific set of semantic labels.
Descriptive statistics of each domain set are shown in Table~\ref{tab:data}.
As we can see, the museum and soccer domains are the only domains which have unknown attributes.
The city domain has many semantic labels and attributes while the museum domain contains more data sources.

\begin{table*}[ht]
  \centering
  \caption{Description of data sources.}\label{tab:data}
		\begin{tabular}{ccccccc} 
		\toprule
		\multirow{2}{*}{\textbf{Domain}} & \# & \# semantic & \# & \# unknown & avg \# rows & avg \# attributes\\
		 & sources & labels & attributes & attributes & per source & per source\\
		\midrule
		weather & 4 & 12 & 44 & 0 & 108.5 & 11 \\
		weapons & 15 & 28 & 175 & 0 & 54.46 & 11.66 \\
		museum & 29 & 20 & 443 & 159 & 6978.89 & 15.27 \\
		soccer & 12 & 18 & 138 & 42 & 2120.16 & 11.5 \\
		city & 10 & 52 & 520 & 0 & 2251 & 52 \\
		\bottomrule
		\end{tabular} 
\end{table*}

\begin{figure*}[ht]\small
\pgfplotsset{
    small,
    legend style={
        at={(0.01,0.01)},
        anchor=south west,
    },
   }%
\begin{minipage}[b]{.3\linewidth}
	\centering
	\begin{tikzpicture}[scale=.8]
	\centering
	\begin{axis}[
			    xlabel={numeric identifiers of semantic labels},
			    ylabel={number of attributes},
			    xmin=0, xmax=12,
			    ymin=0, ymax=5,
			    xtick={1,12},
			    ytick={0,1,2,3,4},
			    legend pos=north east,
			    ymajorgrids=true,
			    grid style=dashed,
			    ybar
			]

			\addplot[
			    color=blue,
			    mark=*
			    ]
			    coordinates {
			    (1,4)(2,4)(3,4)(4,4)(5,4)(6,4)(7,4)(8,4)(9,4)(10,4)(11,3)(12,1)
			    };
			\end{axis}
	\end{tikzpicture}
	\subcaption{weather}
\end{minipage}%
\qquad%
\begin{minipage}[b]{.3\linewidth}
	\centering
	\begin{tikzpicture}[scale=.8]
	\centering
	\begin{axis}[
			    xlabel={numeric identifiers of semantic labels},
			    xmin=0, xmax=28,
			    ymin=0, ymax=30,
			    xtick={1,12,28},
			    ytick={0,5,10,15,20,25,30},
			    legend pos=north east,
			    ymajorgrids=true,
			    grid style=dashed,
			    ybar
			]

			\addplot[
			    color=red,
			    mark=*
			    ]
			    coordinates {
			    (1,26)(2,18)(3,17)(4,15)(5,12)(6,12)(7,11)(8,10)(9,9)(10,8)(11,6)(12,5)(13,4)(14,3)(15,3)(16,3)(17,2)(18,1)(19,1)(20,1)(21,1)(22,1)(23,1)(24,1)(25,1)(26,1)(27,1)(28,1)
			    };
			\end{axis}
	\end{tikzpicture}
	\subcaption{weapons}
\end{minipage}%
\qquad%
\begin{minipage}[b]{.3\linewidth}
	\centering
	\begin{tikzpicture}[scale=.8]
	\centering
	\begin{axis}[
			    xlabel={numeric identifiers of semantic labels},
			    xmin=0, xmax=20,
			    ymin=0, ymax=45,
			    xtick={1,10,19},
			    ytick={0,10,20,30,40},
			    legend pos=north east,
			    ymajorgrids=true,
			    grid style=dashed,
			    ybar
			]

			\addplot[
		    color=green,
		    mark=*,
		    ]
		    coordinates {
		    (1,42)(2,14)(3,11)(4,10)(5,9)(6,9)(7,8)(8,8)(9,5)(10,5)(11,3)(12,3)(13,3)(14,2)(15,2)(16,1)(17,1)(18,1)(19,1)
		    };
			\end{axis}
	\end{tikzpicture}
	\subcaption{soccer}
\end{minipage}
\qquad%
\begin{minipage}[b]{.3\linewidth}
	\centering
	\begin{tikzpicture}[scale=.8]
	\centering
	\begin{axis}[
			    xlabel={numeric identifiers of semantic labels},
			    ylabel={number of attributes},
			    xmin=0, xmax=21,
			    ymin=0, ymax=160,
			    xtick={1,10,21},
			    ytick={0,20,40,60,80,100,120,140,159},
			    legend pos=north east,
			    ymajorgrids=true,
			    grid style=dashed,
			    ybar
			]

			\addplot[
		    color=brown,
		    mark=*
		    ]
		    coordinates {
		    (1,159)(2,31)(3,28)(4,26)(5,23)(6,22)(7,22)(8,21)(9,20)(10,19)(11,16)(12,15)(13,7)(14,7)(15,7)(16,6)(17,5)(18,4)(19,2)(20,2)(21,1)
		    };
			\end{axis}
	\end{tikzpicture}
	\subcaption{museum}
\end{minipage}%
\qquad%
\begin{minipage}[b]{.3\linewidth}
	\centering
	\begin{tikzpicture}[scale=.8]
	\centering
	\begin{axis}[
			    xlabel={numeric identifiers of semantic labels},
			    xmin=0, xmax=52,
			    ymin=0, ymax=15,
			    xtick={1,10,20,30,40,52},
			    ytick={0,10},
			    legend pos=north east,
			    ymajorgrids=true,
			    grid style=dashed,
			    ybar
			]
			\addplot[
		    color=violet,
		    mark=*
		    ]
		    coordinates {
		    (1,10)(2,10)(3,10)(4,10)(5,10)(6,10)(7,10)(8,10)(9,10)(10,10)(11,10)(12,10)(13,10)(14,10)(15,10)(16,10)(17,10)(18,10)(19,10)(20,10)(21,10)(22,10)(23,10)(24,10)(25,10)(26,10)(27,10)(28,10)(29,10)(30,10)(31,10)(32,10)(33,10)(34,10)(35,10)(36,10)(37,10)(38,10)(39,10)(40,10)(41,10)(42,10)(43,10)(44,10)(45,10)(46,10)(47,10)(48,10)(49,10)(50,10)(51,10)(52,10)
		    };
			\end{axis}
	\end{tikzpicture}
	\subcaption{city}
\end{minipage}
\caption{Distribution of attributes according to semantic labels, including the unknown class, in different domains. 
We can see class imbalance in the museum, soccer and weapons domains.
On the $x$-axis we have semantic labels sorted by the number of attributes in each class. 
The $y$-axis shows the number of attributes.}\label{fig:distro}
\end{figure*}
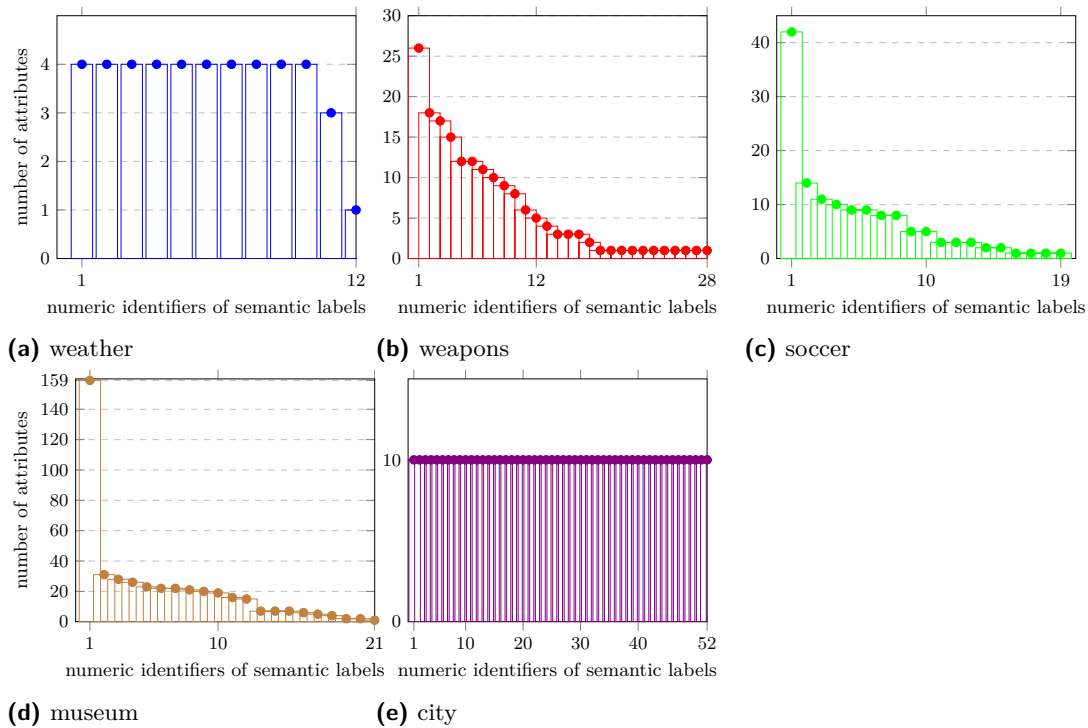

To estimate class imbalance within each domain, we plot the class distribution in Figure~\ref{fig:distro}.
The museum domain has the highest imbalance among classes, the soccer and weapons domains also have imbalanced classes, whereas the weather and city domains have equally represented classes.

\subsection{Experimental setting}

We establish a common evaluation framework for the approaches as described in Section~\ref{sec:approach}.
As a performance metric we use Mean Reciprocal Rank (MRR)~\cite{Craswell:Mean}.
To derive a comprehensive estimate of performance within domains, we implement two cross-validation techniques: leave one out and repeated holdout.

The leave one out strategy is defined as using one source as the testing sample and the rest of sources in the domain as the training samples.
This procedure is repeated as many times as there are sources in the domain.
We calculate MRR on the testing sample and report the average MRR as the final performance metric for each iteration.
For example, for the domain museum we obtain 29 models in total where each model is trained on a different 28 sources, and MRR is calculated on the prediction outcome for a single source.
This strategy allows us to estimate the performance of the different models given that there are enough instances per each semantic label.

In repeated holdout strategy, we randomly sample a ratio $p$ of sources to place in the training sample and use the remaining sources for testing sample, and this procedure is repeated $n$ times.
The final MRR score is an average of MRR scores in each iteration.
We use this technique to simulate the scenario when there is a shortage of labeled sources.
We set the ratio $p=0.2$ and the number of iterations $n=10$.

\subsection{Results}

In this section we report the results of our experiments.
In total we evaluate 13 models, and we report run times required to train the considered models.

\begin{table*}[ht]\small
  \centering
  \caption{MRR scores for leave one out strategy when unknown attributes are not considered.}\label{tab:mrrloo}
  	\begin{tabular}{cllccccc}
  		\toprule
  		{Sampling} & \multicolumn{2}{c}{Model} & city & museum & soccer & weapons & weather\\
  		\midrule
  		\multirow{4}{*}{None} & \multicolumn{2}{c}{DSL} & $0.711$ & $0.848$ & $0.865$ & $0.731$ & $0.904$\\
  		& DINT & all &  \textbf{0.971} & 0.902 & 0.801 & 0.826 & 0.964\\
  		& DINT & base &  0.925 & 	0.86 & 0.794 & 0.810 & 0.93
\\
  		& DINT & base+ &  0.925 & 0.89 & 0.792 & 0.807 & 0.93
\\
  		\midrule
		\multirow{5}{*}{Bagging} & \multicolumn{2}{c}{MLP} & $0.873$ & $0.886$ & $0.862$ & $0.799$ & 0.965 \\
		& \multicolumn{2}{c}{CNN} &  $0.877$ & $0.893$ & $0.823$ & $0.813$ & $0.939$
 \\
		& DINT & all & 0.956 & \textbf{0.913} & 0.804 & \textbf{0.833} & \textbf{0.979} \\
  		& DINT & base & 0.928 & 0.894 & \textbf{0.887} & 0.825 & 0.941\\
  		& DINT & base+ & 0.928 & 0.911 & 0.79 & 0.813 & 0.956\\
  		\midrule
		Resample & DINT & all & 0.969 & 0.907 & 0.814 & 0.792 & 0.956\\
  		To & DINT & base & $0.929$ & 0.888 & 0.802 & 0.788 & 0.911\\
  		Mean & DINT & base+ & 0.929 & 0.901 & 0.835 & 0.765 & 0.926 \\
		\bottomrule
	\end{tabular}
\end{table*}

\begin{table*}[ht]
  \centering
  \caption{Model training times (s) for leave one out.}\label{tab:timeloo}
  	\begin{tabular}{cllccccc}
  		\toprule
  		{Sampling} & \multicolumn{2}{c}{Model} & city & museum & soccer & weapons & weather\\
  		\midrule
  		\multirow{4}{*}{None} & \multicolumn{2}{c}{DSL} &  295.6 & 164.3 &	36.6 & 269.5 & 8.2\\
  		& DINT & all & 10.8 & 74.8 & 8.0 & 6.2 & 2.0\\
  		& DINT & base &  \textbf{10.2} & 20.4 & 3.9 & 5.0 & 3.6\\
  		& DINT & base+ & \textbf{10.2} & 20.7 & 4.0 & \textbf{4.1} & \textbf{2.0}\\
  		\midrule
		\multirow{5}{*}{Bagging} & \multicolumn{2}{c}{MLP} & $184.2$ & $216.5$ & $26.8$ & $85$ & $11$ \\
		& \multicolumn{2}{c}{CNN} &  $184.8$ & $276.2$ & $29.7$ & $71.3$ & $12.5$\\
		& DINT & all &  $212$ & $310.3$ & $47.3$ & $127.9$ & $11.6$ \\
  		& DINT & base & 165.5 & 83.1 & 26.8 & 35.1 & 8.6 \\
  		& DINT & base+ & 165.5 & 80.0 & 27.8 & 44.0 & 7.1\\
  		\midrule
		Resample & DINT & all & 10.9 &  58.6 & 12.2 & 8.3	& 2.1\\
  		To & DINT & base &  10.8 & 20.5 & \textbf{2.3}	& 4.3	& 2.1 \\
  		Mean & DINT & base+ &  10.9 & \textbf{18.4} & 2.8	& 4.3	& 2.1\\
		\bottomrule
	\end{tabular}
\end{table*}

To train \emph{MLP} and \emph{CNN} models, we need many training instances, so we use bagging (presented in Section~\ref{sec:bagging}) with parameters \emph{numBags}=150 and \emph{bagSize}=100 to increase the size of the initial training set.
We can train the semantic labeling system \emph{DINT} with different sampling strategies.
In particular, we report results when we apply no resampling and bagging with parameters \emph{bagSize}=100 and \emph{numBags}=100.
We also experiment with various class imbalance resampling strategies, including resampling to the mean or maximum of instance counts per class. For brevity and without loss of generality we report results only for the resampling to mean strategy denoted as \emph{ResampleToMean}. By design \emph{DSL} and \emph{DSL+} use no resampling.

As mentioned in Section~\ref{sec:dint}, the \emph{DINT} model is built on a set of elaborately engineered features.
\emph{MLP} model, on the other hand, uses only \emph{chardist} and \emph{entropy}.
To better compare the performance of \emph{MLP} and \emph{DINT}, we create a new model \emph{DINT base} and reduce the number of features to just \emph{chardist} and \emph{entropy}.
In addition, we create another model \emph{DINT base+} by using \emph{chardist} and \emph{entropy} and add a feature \emph{minimum edit distance}. We choose this feature as feature importance scores produced by the random forest algorithm rank edit distance higher than the other features extracted from names.

Table~\ref{tab:mrrloo} reports the MRR scores for leave one out strategy.
Surprisingly, models built on just normalized character distributions of attribute values perform in many cases very well.
Deep learning models \emph{MLP} and \emph{CNN} are often comparable with \emph{DINT} models, however they come usually at a higher computational cost.
Run times for training each model are shown in Table~\ref{tab:timeloo}.

As we can see, \emph{DINT} models that use bagging to sample more training instances achieve the best results in four domains.
Remarkably, these are also the domains with higher class imbalance and variety among data sources in terms of number of rows and number of columns.
Data sources in the city domains have the same number of attributes.
We have also discovered that bagging needs to be performed both at the training and prediction stages to achieve the best performance.
We have observed that this setting makes a noticeable difference in domains where the number of rows varies substantially among data sources.
For example, in the museum domain number of rows ranges from 6 to 85235, and in the soccer domain the range is from 500 to 9443.

In terms of computation time, the best performing model \emph{DINT all} for the museum domain requires a lot of time for training. 
The most computationally expensive features are four different edit distances: minimum edit distance, JCN, LIN and $k$-nearest neighbors.
This suggests that the \emph{DINT} model with all possible features does not scale well with the increasing number of attributes in the training set.
Considering similarity metrics used in other approaches like \emph{DSL} and \emph{T2K}~\cite{Ritze:matching}, computing TF-IDF and Jaccard's scores may help resolve this runtime issue for \emph{DINT all}.

For class imbalance, although the \emph{ResampleToMean} strategy improves the performance of \emph{DINT} models with no sampling in the domains with the highest class imbalance (i.e., museum and soccer), it appears that the \emph{ResampleToMean} strategy leads to a decreased performance in the domains with a less prominent imbalance (i.e., weapons and weather).
This leads us to the idea that a class resampling strategy needs to be improved.

One potential strategy may be in combining bagging and resampling strategies.
Instead of fixing \emph{numBags} for all attributes, the parameter could be changed to be either the mean or maximum of instance counts per each class.
In such a way we can perform a resampling strategy which does not produce replicas of the attributes.

Apart from the city and weapons domains, our newly designed models have a similar performance to \emph{DSL}.
However, the computational complexity of these models varies.
For the museum domain \emph{DINT base+} has a higher MRR than \emph{DSL}, yet \emph{DINT base+} needs half the time less for training.
It appears that attributes which contain a mixture of textual and numeric are a bottleneck for \emph{DSL} since data sources in the city and weapons domains have multiple mixed data columns.

\begin{table*}[ht]\small
  \centering
  \caption{MRR scores for repeated holdout strategy when unknown attributes are not considered.}\label{tab:mrrrep}
  	\begin{tabular}{cllccccc} 
  		\toprule
  		{Sampling} & \multicolumn{2}{c}{Model} & city & museum & soccer & weapons & weather\\
  		\midrule
  		\multirow{5}{*}{None} & \multicolumn{2}{c}{DSL} & 0.719 & 0.889 & 0.614 & 0.611 & 0.805\\
  		 & \multicolumn{2}{c}{DSL+} & 0.782 & 	\textbf{0.927} & \textbf{0.813} & \textbf{0.872} & \textbf{1}\\
  		& DINT & all & \textbf{0.949} & 0.798 & 0.553 & 0.688 & 0.583\\
  		& DINT & base &  0.888 & 0.763 & 0.516 & 0.684 & 0.621\\
  		& DINT & base+ & 0.888 & 0.778 & 0.542 & 0.686 & 0.621\\
  		\midrule
		\multirow{5}{*}{Bagging} & \multicolumn{2}{c}{MLP} & $0.797$ & $0.77$ & $0.663$ & $0.695$ & $0.887$\\ 
		& \multicolumn{2}{c}{CNN} &  $0.723$ & $0.774$ & $0.606$ & $0.664$ & $0.882$\\ 
		& DINT & all &  0.945 & 0.791 & 0.656 & 0.682 & 0.854\\
  		& DINT & base & 0.919 & 0.788 & 0.634 & 0.701 & 0.867\\
  		& DINT & base+ &  0.919 & 0.790 & 0.628 & 0.688 & 0.852\\
  		\midrule
		Resample & DINT & all & \textbf{0.949} & 0.789 & 0.455 & 0.588 & 0.557\\
  		To & DINT & base & 0.89 & 0.749 & 0.451 & 0.578 & 0.611\\
  		Mean & DINT & base+ & 0.89 & 0.758 & 0.445 & 0.564 & 0.611\\
		\bottomrule
	\end{tabular} 
\end{table*}

In cases where there are few labeled instances (repeated holdout strategy in Table~\ref{tab:mrrrep}), we observe that \emph{DSL} performs well, especially \emph{DSL+}, which leverages labeled instances from other domains.
We should be aware that in this scenario there are many unseen labels, which makes MRR ill-defined.
If we compare \emph{DINT} models in this scenario, it suggests that bagging is advantageous in situations when there are few labeled attributes.
Overall, enhancing our \emph{DINT} model, which uses simple features and bagging, with \emph{DSL+} knowledge transfer capability might result in a more stable semantic labeling system. 
Another enhancement may be to introduce resampling strategies into the \emph{DSL} system.

In addition, we perform experiments for the two domains museum and soccer, where unmapped attributes cause skewed class distributions. 
Here we want to establish how well different approaches can recognize such attributes.
In Tables~\ref{tab:unknownloo} and~\ref{tab:unknownrep} we can see that the performance of semantic labeling systems changes considerably.
Both the \emph{DSL} and \emph{DSL+} performance is affected by their inability to differentiate "unwanted" attributes.

When performing bagging on attributes in the training data, we introduce diversity by drawing many samples of attribute values.
However, we do not apply any perturbation technique to the names of the attributes and instead use their exact replicas.
In Table~\ref{tab:unknownrep} we observe that \emph{DINT base} performs better than \emph{DINT base+} when bagging is used.
In datasets with scarce labeled instances our DINT models tend to overfit the attribute names that are present in the training data.
This suggests that introducing a technique similar to bagging for column headers might lead to a much better performance.
On the other hand, our results are consistent with the observations in the work of Ritze et al.\cite{Ritze:matching}.
Their results indicate that comparing attribute values is crucial for this task while attribute names might introduce additional noise.

\begin{table*}[t]\small
  \centering
  \caption{Performance for leave one out strategy when unknown class is considered.}
  	\label{tab:unknownloo}
  	\begin{tabular}{cll|cc|cc} 
  		\toprule
  		\multirow{2}{*}{Sampling} & \multicolumn{2}{c|}{\multirow{2}{*}{Model}} & \multicolumn{2}{c|}{MRR scores} & \multicolumn{2}{c}{Train time ($s$)}\\
  		 & & & museum & soccer & museum & soccer\\
  		\midrule
  		\multirow{4}{*}{None} & \multicolumn{2}{c|}{DSL} & $0.56$ & $0.618$ & $156.6$ & $36.3$\\
  		& DINT & all & 0.866 & $0.827$ & $100.6$ & $6.8$\\
  		& DINT & base & $0.838$ & $0.809$ & $28.4$ & $5.9$\\
  		& DINT & base+ & $0.849$ & $0.824$ & $33.4$ & $6.2$\\
  		\midrule
		\multirow{5}{*}{Bagging} & \multicolumn{2}{c|}{MLP} & $0.802$ & $0.784$& $417.2$ & $37.6$\\ 
		& \multicolumn{2}{c|}{CNN} & $0.831$ & $0.785$ &  $394.5$ & $39.6$\\ 
		& DINT & all & 0.854 & 0.795 & 395.2 	  & 64.5\\
  		& DINT & base & $0.839$ & \textbf{0.863} & $112.5$ & $26.7$\\
  		& DINT & base+ & \textbf{0.867} & $0.793$ & $114.4$ & $30.6$\\
  		\midrule
		Resample & DINT & all  & 0.776 & 	0.730 &  100.5	& 6.8\\
  		To & DINT & base & $0.721$ & $0.69$& \textbf{26.2} & \textbf{4.2}\\
  		Mean & DINT & base+ & $0.759$ & $0.753$& $26.7$ & $5.2$\\
		\bottomrule		
		\end{tabular} 
\end{table*}

\begin{table*}[t]\small
  \centering
  \caption{Performance for repeated holdout strategy when unknown class is considered.}
  	\label{tab:unknownrep}
  	\begin{tabular}{cll|cc|cc}
  		\toprule
  		\multirow{2}{*}{Sampling} & \multicolumn{2}{c|}{\multirow{2}{*}{Model}} & \multicolumn{2}{c|}{MRR scores} & \multicolumn{2}{c}{Train time ($s$)}\\
  		 & & & museum & soccer & museum & soccer\\
  		\midrule
  		\multirow{5}{*}{None} & \multicolumn{2}{c|}{DSL} & $0.544$ & $0.355$ &  $15.5$ & $4.2$\\
  		 & \multicolumn{2}{c|}{DSL+} &  $0.303$ & $0.43$ &  $215.9$ & $241.3$\\
  		& DINT & all & 0.769 & $0.549$ & $16.2$ & $2$\\
  		& DINT & base  & $0.743$ & $0.608$& $10.2$ & $2$\\
  		& DINT & base+  & $0.742$ & \textbf{0.613} & \textbf{10.1} & \textbf{2}\\
  		\midrule
		\multirow{5}{*}{Bagging} & \multicolumn{2}{c|}{MLP} & $0.675$ & $0.572$ & $94.4$ & $9.4$\\
		& \multicolumn{2}{c|}{CNN}  & $0.683$ & $0.534$ & $87.7$ & $10$\\
		& DINT & all  &  \textbf{0.827} & 0.551 & 101.9 & 13.6\\
  		& DINT & base  & $0.76$ & $0.593$ & $55.6$ & $13.5$\\
  		& DINT & base+  & $0.721$ & $0.59$ &$53.1$ & $13.6$\\
  		\midrule
		Resample & DINT & all  & 0.637	&0.428 & 35.5 & 2.6\\
  		To & DINT & base  & $0.607$ & $0.475$  &$12.2$ & $2$\\
  		Mean & DINT & base+  & $0.633$ & $0.479$& $11.5$ & $2$\\
		\bottomrule
		\end{tabular}
\end{table*}

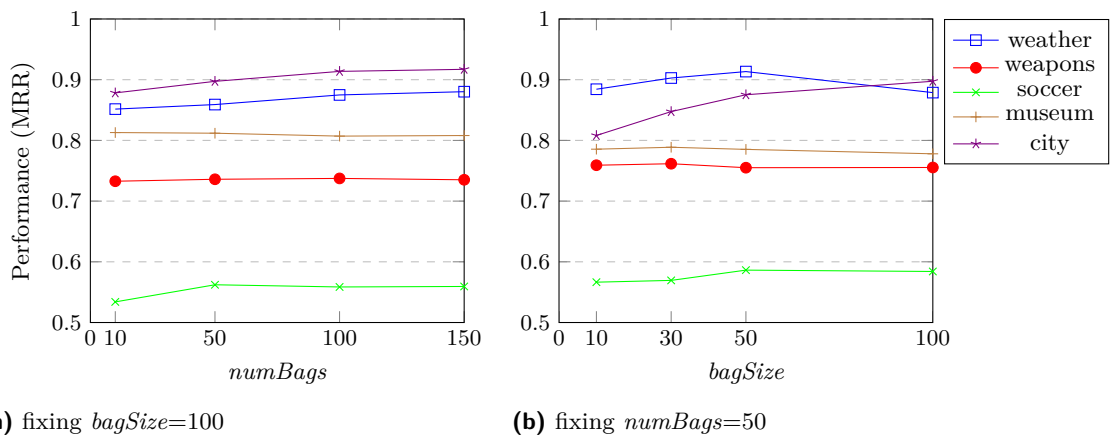
\begin{figure*}[ht]\small
\pgfplotsset{
    small,
    legend style={
        at={(0.01,0.01)},
        anchor=south west,
    },
   }%
\begin{minipage}[b]{.5\linewidth}
\centering
\begin{tikzpicture}[baseline]
\begin{axis}[
		    xlabel={\emph{numBags}},
		    ylabel={Performance (MRR)},
		    xmin=0, xmax=150,
		    ymin=0.5, ymax=1,
		    xtick={0,10,50,100,150},
		    ytick={0.5,0.6,0.7,0.8,0.9,1},
		    legend pos=south east,
		    ymajorgrids=true,
		    grid style=dashed
		]

		\addplot[
		    color=blue,
		    mark=square,
		    ]
		    coordinates {
		    (10, 0.85154623154623166)(50, 0.85902097902097896)(100,0.87487956487956486)(150,0.8802331002331002)
		    };
		\addplot[
		    color=red,
		    mark=*,
		    ]
		    coordinates {
		    (10,  0.73272601851632158)(50, 0.73596072887032105)(100,0.73736629035127532)(150,0.73512315058140676)
		    };
		\addplot[
		    color=green,
		    mark=x,
		    ]
		    coordinates {
		    (10,  0.53396973928187841)(50, 0.56218126081141884)(100,0.55857468339226135)(150,0.55938671902457904)
		    };
		 \addplot[
		    color=brown,
		    mark=+,
		    ]
		    coordinates {
		    (10,  0.81292733659634253)(50, 0.81182668838989458)(100, 0.80705336361770219)(150,0.80801980454223066)
		    };
		 \addplot[
		    color=violet,
		    mark=star,
		    ]
		    coordinates {
		    (10,   0.87823326846764416)(50, 0.89727564102564272)(100,0.9136698717948738)(150,0.91708333333333425)
		    };

		\end{axis}
\end{tikzpicture}
\subcaption{fixing \emph{bagSize}=100}\label{fig:numbags}
\end{minipage}%
\begin{minipage}[b]{.5\linewidth}
\centering
\begin{tikzpicture}[baseline]
		\begin{axis}[
		    xlabel={\emph{bagSize}},
		    xmin=0, xmax=100,
		    ymin=0.5, ymax=1,
		    xtick={0,10,30,50,100},
		    ytick={0.5,0.6,0.7,0.8,0.9,1},
		    legend pos=outer north east,
		    ymajorgrids=true,
		    grid style=dashed
		]

		\addplot[
		    color=blue,
		    mark=square,
		    ]
		    coordinates {
		    (10, 0.88427350427350415)(30,  0.9027583527583527)(50,0.91336441336441332)(100,0.87871794871794873)
		    };
		\addplot[
		    color=red,
		    mark=*,
		    ]
		    coordinates {
		    (10,  0.75910921069769555)(30, 0.76156533626682754)(50,0.75502218858295522)(100,0.75538165183154382)
		    };
		\addplot[
		    color=green,
		    mark=x,
		    ]
		    coordinates {
		    (10,  0.56646107828012637)(30, 0.5693901714881866)(50,0.58629781920572777)(100,0.58408351395802571)
		    };
		 \addplot[
		    color=brown,
		    mark=+,
		    ]
		    coordinates {
		    (10,  0.78552344907839267)(30, 0.78870711278952932)(50,0.78519302797324753)(100,0.77792094596490047)
		    };
		 \addplot[
		    color=violet,
		    mark=star,
		    ]
		    coordinates {
		    (10,  0.80810421349483996)(30, 0.84745259081196467)(50,0.8751719894688631)(100,0.89739411630036825)
		    };
		\legend{weather,weapons,soccer,museum,city}

		\end{axis}
	\end{tikzpicture}
\subcaption{fixing \emph{numBags}=50}\label{fig:bagsize}
\end{minipage}
\caption{Dependence of MRR scores for \emph{DINT base} on the bagging parameters using repeated holdout strategy. Unknown attributes are not considered.}\label{fig:bagging}
\end{figure*}

Clearly, the performance of our approach \emph{DINT} varies depending on the chosen bagging parameters \emph{numBags} and \emph{bagSize}.
To explore this dependence, we evaluate the performance of \emph{DINT} with only \emph{chardist} and \emph{entropy} features by varying one of the bagging parameters and fixing the other one.
We report the results of our evaluation in Figure~\ref{fig:bagging}.
Here we do not consider unknown attributes and choose the repeated holdout strategy to analyze the behavior of bagging when there is a shortage of training data.
Interestingly, increasing the values of the bagging parameters does not always lead to an improved performance, though the computational time required for both the training and prediction stages increases.
The city domain is the most sensitive to bagging parameters.
We assume this is because the city domain is the only domain with an equal distribution of semantic labels, equal numbers of columns and rows across data sources.
It appears that in other domains, bagging makes models more robust towards variance in these characteristics.

\section{Related Work}\label{section:rel_work}

The problem of semantic labeling, as addressed in this work, can be regarded as the problem of schema matching in the field of data integration~\cite{Bellahsene:smm}.
In the schema matching problem we match elements between the source and target schemata.
In our case elements of the source schema are attributes, and we want to map these attributes to properties in the ontology.
The semantic labeling problem is also known in literature as attribute-to-property matching~\cite{Ritze:HTML,Ritze:matching}.
Indicating semantic correspondences manually might be appropriate if only few data sources need to be integrated, however, it becomes tedious with the growing number of heterogeneous schemata.
Hence, automatic or semi-automatic approaches for schema matching are being actively developed.

From machine learning perspective, we can categorize these approaches into unsupervised techniques which compute various similarity metrics and supervised techniques which build a multi-class classification model.
Unsupervised approaches are used in SemanticTyper~\cite{Ramnandan:Assigning}, T2K~\cite{Ritze:HTML} and its extended version~\cite{Ritze:matching}.
In all these approaches authors design similarity metrics for attribute names and attribute values, yet one substantial difference is whether additional knowledge is used in the computation.
For example, authors in~\cite{Ritze:HTML} and~\cite{Ritze:matching} leverage contextual information from DBpedia.

Among supervised approaches, there are probabilistic graphical models used in the work of Limaye et al.~\cite{Limaye:Annotating} to annotate web tables with entities for cell values, types for attributes and relationships for binary combinations of attributes.
Mulwad et al.~\cite{Mulwad:Semantic} extend this approach by leveraging information from Wikitology Knowledge Base (KB).
The problem with probabilistic graphical models is though that they do not scale with the number of semantic labels in the domain.
Also, Mulwad et al.\ as well as Venetis et al.~\cite{Venetis:Recovering}, who used the isA database KB, extract additional data from knowledge bases to assign a semantic label to an attribute.
Hence, these approaches are limited to domains well represented in those knowledge bases.
Our approach, on the other hand, is not domain specific and allows a model to be trained on any data.
However, we cannot apply a model learnt on one domain to another, which is possible with the DSL approach~\cite{Pham:semantic}.

To the best of our knowledge, DSL introduced by Pham et al.\cite{Pham:semantic} is among the top semantic labeling systems.
Pham et al.\ compare DSL to their previous approach SemanticTyper~\cite{Ramnandan:Assigning} and T2K system~\cite{Ritze:HTML}, and achieve higher MRR scores on a variety of datasets.
Therefore, we use DSL as the state-of-the art model in our benchmark to evaluate our new approaches.

Ritze et al.~\cite{Ritze:matching} and Pham et al.~\cite{Pham:semantic} mention the problem of the unknown class.
In the first work the authors discuss "unwanted" attributes while in the second work the authors reflect on how to handle "unseen" attributes.
In our work we do not differentiate between these two cases and show that we can successfully identify such attributes when sufficient training data is available.

\section{Conclusion}

In this paper we have studied the problem of supervised semantic labeling and have conducted experiments to evaluate how different approaches perform at this task.
Our main finding is that our bagging sampling technique can provide meaningful diversity to our training data to improve performance.
Additionally, this technique can overcome the lack of labeled attributes in the domain and can increase the number of instances for under-represented semantic labels.
We find that given scarce training data, bagging leads to a noticeable improvement in performance, though the state-of-the-art system \emph{DSL}~\cite{Pham:semantic} achieves a better precision by leveraging information about labeled instances from other domains.
However, if we are to consider unwanted attributes and unseen semantic labels, our new system \emph{DINT} demonstrates the best performance.
Among the semantic labeling systems in our benchmark we have observed that the performance results are highly dependent on the use case.

We have also shown that deep learning models, such as \emph{CNN} and \emph{MLP}, can also be applied to solve this problem.
Though these models do not excel in performance in the majority of cases, their advantage is the simplicity of features extracted from attributes.
For example, \emph{CNN} is built on raw sequences of attribute values.
Surprisingly, we have discovered that even random forests constructed just on character distributions of values and entropy of attributes provide remarkable results in many cases.
This supports the observations in literature that attribute values are crucial for semantic labeling task~\cite{Ritze:HTML,Ritze:matching}.

Future work may involve exploring a combination of bagging and class imbalance resampling strategies.
We have observed that where the domain data has high imbalance among representatives of different semantic labels, resampling can lead to an improved performance but a more sophisticated approach is required in domains which do not exhibit these characteristics.
Another possible direction for improvement is to introduce an equivalent of bagging for attribute names.
In addition, our experiments indicate that the performance of systems is often affected by the variance in sizes of data sources and how well each semantic label is represented in the training data.
To this end, we consider including T2KMatch~\cite{Ritze:matching} into our benchmark as well as domain sets from the RODI benchmark~\cite{Pinkel:rodi}.

\bibliography{schemamapping}

\end{document}